\documentclass{article}

\usepackage[dblblindworkshop,final]{neurips_2025}
\workshoptitle{MATH-AI}

\usepackage{amsmath, amssymb, amsfonts, mathtools, amsthm}
\usepackage[utf8]{inputenc}
\usepackage[T1]{fontenc}

\usepackage{booktabs, multirow, array, colortbl}

\usepackage{graphicx, xcolor, wrapfig, subcaption, caption}
\usepackage{tikz}

\usepackage{microtype, url, enumitem, pifont, nicefrac, fontawesome}
\usepackage{cancel}
\usepackage{mdframed}

\usepackage{listings}
\usepackage{algorithm}
\usepackage{algpseudocode}

\usepackage[most,skins,theorems]{tcolorbox}
\tcbset{
  aibox/.style={
    width=\linewidth,
    top=7pt,
    bottom=2pt,
    colback=blue!6!white,
    colframe=black,
    colbacktitle=black,
    enhanced,
    center,
    attach boxed title to top left={yshift=-0.1in,xshift=0.15in},
    boxed title style={boxrule=0pt,colframe=white,},
  }
}
\newtcolorbox{AIbox}[2][]{aibox,title=#2,#1}

\definecolor{rliableolive}{HTML}{BBCC33}
\definecolor{rliableblue}{HTML}{77AADD}
\definecolor{rliablered}{HTML}{EE8866}
\definecolor{SDEblue}{RGB}{28,58,88}
\definecolor{cc1}{rgb}{1.0, 0.44, 0.37}
\definecolor{cc2}{rgb}{0.0, 0.2, 0.6}
\definecolor{cc3}{RGB}{255, 191, 0}
\definecolor{cc4}{RGB}{0, 128, 128}

\theoremstyle{definition}

\usepackage{hyperref}
\usepackage{cleveref}

\crefname{section}{Sec.}{Sec.}
\crefname{theorem}{Theorem}{Theorems}
\crefname{corollary}{Corollary}{Corollaries}
\crefname{lemma}{Lemma}{Lemmas}
\crefname{equation}{Eq.}{Eq.}
\crefname{proposition}{Proposition}{Propositions}
\crefname{claim}{Claim}{Claims}
\crefname{remark}{Remark}{Remarks}
\crefname{observation}{Observation}{Observations}
\crefname{assumption}{Assumption}{Assumptions}
\crefname{template}{Template}{Templates}
\crefname{definition}{Definition}{Definitions}
\crefname{appendix}{App.}{Apps.}
\crefname{algorithm}{Algorithm}{Algorithms}
\crefname{figure}{Fig.}{Fig.}
\crefname{table}{Table}{Tables}
\crefname{property}{Property}{Properties}
\crefname{line}{Line}{Lines}

\definecolor{table-blue}{RGB}{173, 216, 230}
\definecolor{darkblue}{rgb}{0, 0, 0.5}
\hypersetup{
  colorlinks=true,
  citecolor=darkblue,
  linkcolor=darkblue,
  urlcolor=magenta
}

\title{Understanding R1-Zero-Like Training: A Critical Perspective}

\newcommand{\sail}{1}
\newcommand{\nus}{2}
\newcommand{\smu}{3}
\newcommand{\sailnus}{1,2}
\newcommand{\sailsmu}{1,3}
\renewcommand\footnotemark{}
\author{
\hspace{2.5em}
Zichen Liu\textsuperscript{*$\dagger$\sailnus}\thanks{\llap{$^*$}Core Contributors.}\thanks{\llap{$^\dagger$}Project Lead.}
\quad
Changyu Chen\textsuperscript{*\sailsmu} \quad
Wenjun Li\textsuperscript{*\smu}
\quad
Penghui Qi\textsuperscript{*\sailnus}
\vspace{-1.2em}
\And
\hspace{5em}
Tianyu Pang\textsuperscript{\sail} \quad
Chao Du\textsuperscript{\sail} \quad
Wee Sun Lee\textsuperscript{\nus}\quad 
Min Lin\textsuperscript{\sail} \quad 
\vspace{1em}
\\
\textsuperscript{\sail}Sea AI Lab \quad
\textsuperscript{\nus}National University of Singapore \quad
\textsuperscript{\smu}Singapore Management University
}

\title{You Need Reasoning to Learn Reasoning: The Limitations of Label-Free RL in Weak Base Models
}

\author{
  Shuvendu Roy, Hossein Hajimirsadeghi, Mengyao Zhai, Golnoosh Samei \\
  RBC Borealis \\
  \textit{shuvendu.roy@rbc.com, \{hossein.hajimirsadeghi, mengyao.zhai, golnoosh.samei\}@borealisai.com}
}

\begin{document}

\maketitle

\begin{abstract}
\vspace{-7pt}
Recent advances in large language models have demonstrated the promise of unsupervised reinforcement learning (RL) methods for enhancing reasoning capabilities without external supervision. However, the generalizability of these label-free RL approaches to smaller base models with limited reasoning capabilities remains unexplored. In this work, we systematically investigate the performance of label-free RL methods across different model sizes and reasoning strengths, from 0.5B to 7B parameters. Our empirical analysis reveals critical limitations: label-free RL is highly dependent on the base model's pre-existing reasoning capability, with performance often degrading below baseline levels for weaker models. We find that smaller models fail to generate sufficiently long or diverse chain-of-thought reasoning to enable effective self-reflection, and that training data difficulty plays a crucial role in determining success. To address these challenges, we propose a simple yet effective method for label-free RL that utilizes curriculum learning to progressively introduce harder problems during training and mask no-majority rollouts during training. Additionally, we introduce a data curation pipeline to generate samples with predefined difficulty. Our approach demonstrates consistent improvements across all model sizes and reasoning capabilities, providing a path toward more robust unsupervised RL that can bootstrap reasoning abilities in resource-constrained models. We make our code available at \href{https://github.com/BorealisAI/CuMa}{https://github.com/BorealisAI/CuMa}.
\end{abstract}

\vspace{-14pt}
\section{Introduction}
\vspace{-7pt}
Recent advances in large language models (LLMs) have highlighted the effectiveness of reinforcement learning (RL) techniques for enhancing reasoning capabilities, particularly in domains like mathematics and code generation. However, traditional approaches such as Reinforcement Learning from Human Feedback (RLHF) and Reinforcement Learning with Verifiable Rewards (RLVR) rely heavily on external supervision, including human annotations or domain-specific ground-truth labels \cite{shao2024deepseekmath, guo2025deepseek}. This dependency poses significant scalability challenges, as acquiring such supervision becomes increasingly costly and impractical for emerging, complex tasks. To address this, recent works have proposed unsupervised paradigms that enable models to self-improve without labelled data. For instance, Test-Time Reinforcement Learning (TTRL) \cite{zuo2025ttrl} leverages majority voting on unlabeled test data to estimate rewards, allowing models to adapt and evolve during inference. Similarly, Reinforcement Learning from Internal Feedback (RLIF), as exemplified by Intuitor \cite{zhao2025intuitor}, uses intrinsic signals like the model's own confidence (self-certainty) to drive optimization, eliminating the need for external verifiers.

Despite these promising developments, existing unsupervised RL approaches have primarily been evaluated on relatively large encoder-only models that already possess decent reasoning capabilities. For example, both TTRL and Intuitor focus on backbones from the Qwen series, such as Qwen2.5-Math-7B and Qwen2.5-7B, which are known for their strong baseline performance in reasoning tasks due to extensive pre-training. However, it remains unclear how these methods perform on models lacking such inherent capabilities, such as smaller LLMs or tasks where the pre-trained LLM does not have pre-existing knowledge for complex reasoning. 

\begin{wrapfigure}{r}{0.5\linewidth}
\vspace{-2pt}
    \centering
    \includegraphics[width=0.95\linewidth]{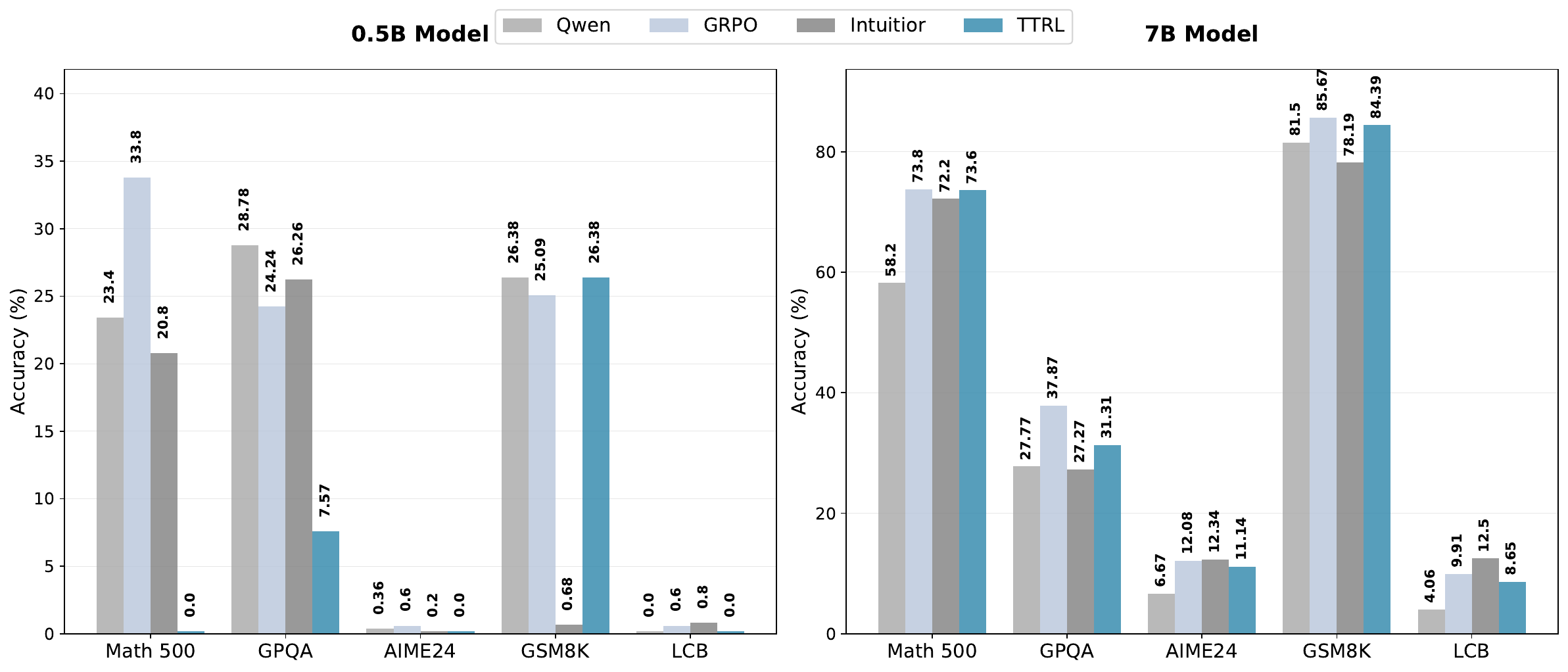}
    \caption{The performance of the Qwen2.5-0.5B base model, compared to the Qwen2.5-7B, shows that smaller models with weaker reasoning capabilities do not improve with label-free RL training.}
    \label{fig:bannar}
    \vspace{-5pt}
\end{wrapfigure}
In this paper, we investigate how unsupervised reinforcement learning methods adapt to smaller, pure base models in a label-free environment. Our findings reveal that these smaller models, which have weaker reasoning capabilities, struggle to learn in unlabeled settings, including both pseudo-labelling and self-consistency setups. We identify several key factors contributing to the failure of existing methods and propose a curriculum-based unsupervised RL approach. Our method demonstrates stronger generalization across different model types and sizes. Below is a summary of our main takeaways.

\vspace{-0.1cm}
\begin{AIbox}{Overview of takeaways}
\begin{itemize}[leftmargin=0em]
\item Label-free RL is highly dependent on the \textbf{reasoning capability of the base model}. Performance drops significantly, sometimes worse than the base model, if the base model’s reasoning ability is insufficient. 
\item Smaller base model (with limited reasoning) \textbf{does not generate a longer chain-of-thought} to elicit self-reflection (Aha moment).
\item Length of chain-of-thought is \textbf{not a direct reflection of strong reasoning} for label-free RL.
\item The \textbf{difficulty of the training data plays an important role}. A base model with limited reasoning ability cannot effectively learn from very hard problems, where it can rarely, if ever, generate the correct solution.
\item Our simple training modification, which employs a \textbf{curriculum learning approach} that begins with easier problems and gradually introduces more challenging ones, enhances the performance of label-free RL across all model sizes and reasoning capabilities. Performance can be further enhanced by curating supplementary training datasets of predefined difficulty levels and employing a masked reward strategy for non-majority samples.
\end{itemize}
\end{AIbox}

\section{Analysis on Label-free RL}
\vspace{-7pt}
\begin{wrapfigure}{r}{0.3\linewidth}
\vspace{-35pt}
    \centering
    \includegraphics[width=0.95\linewidth]{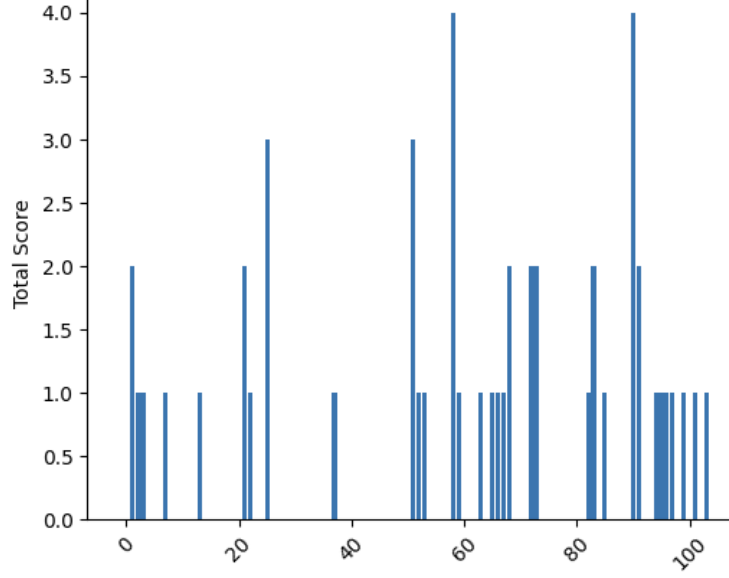}
    \caption{Correct answers generated in early stages of training by Qwen2.5-0.5B.}
    \label{fig:rollouts}
    \vspace{-5pt}
\end{wrapfigure}
\subsection{Preliminary}
\vspace{-5pt}
To establish a foundation for our analysis, we first provide an overview of the two key unsupervised RL methods: Test-Time Reinforcement Learning (TTRL) and Intuitor.
TTRL \cite{zuo2025ttrl} introduces a framework for performing RL directly on unlabeled test data during inference. Given a prompt $x$, the method samples multiple outputs $\{y_1, \dots, y_N\}$ from the policy $\pi_\theta$. A label $y^*$ is estimated via majority voting on the extracted answers, and binary rewards are assigned: $r(\hat{y}_i, y^*) = 1$ 
if $\hat{y}_i = y^*$, else 0. The policy is then optimized to maximize the expected reward using algorithms like Group Relative Policy Optimization (GRPO). Intuitor \cite{zhao2025intuitor} replaces external rewards with the model's intrinsic self-certainty—a measure of confidence defined as the average KL divergence from a uniform distribution over the vocabulary. Using GRPO, advantages are computed from normalized self-certainty scores, guiding the policy toward higher-confidence outputs.

\vspace{-7pt}
\subsection{Label-free RL struggles with smaller models}
\vspace{-7pt}
To investigate the effectiveness of verifier-free reinforcement learning methods on models with varying baseline capabilities, we conduct extensive experiments comparing the performance of Qwen2.5-0.5B and Qwen2.5-7B across multiple reasoning benchmarks. Our results, presented in Figure \ref{fig:bannar}, reveal a striking disparity between the smaller and larger models. For the Qwen2.5-7B model, we observe consistent improvements across all evaluated benchmarks when applying verifier-free RL methods. On Math 500, the base model achieves a score of 58.2 points, which 
increases to 74.6 with TTRL and 72.2 with Intuitor, comparable to the 73.8 achieved by verifier-based (using label) GRPO. Similar positive trends are evident on GPQA, AIME24, GSM8K, and LCB, 
\begin{wrapfigure}{r}{0.3\linewidth}   
    \vspace{-5pt}
    \centering
        \includegraphics[width=\linewidth]{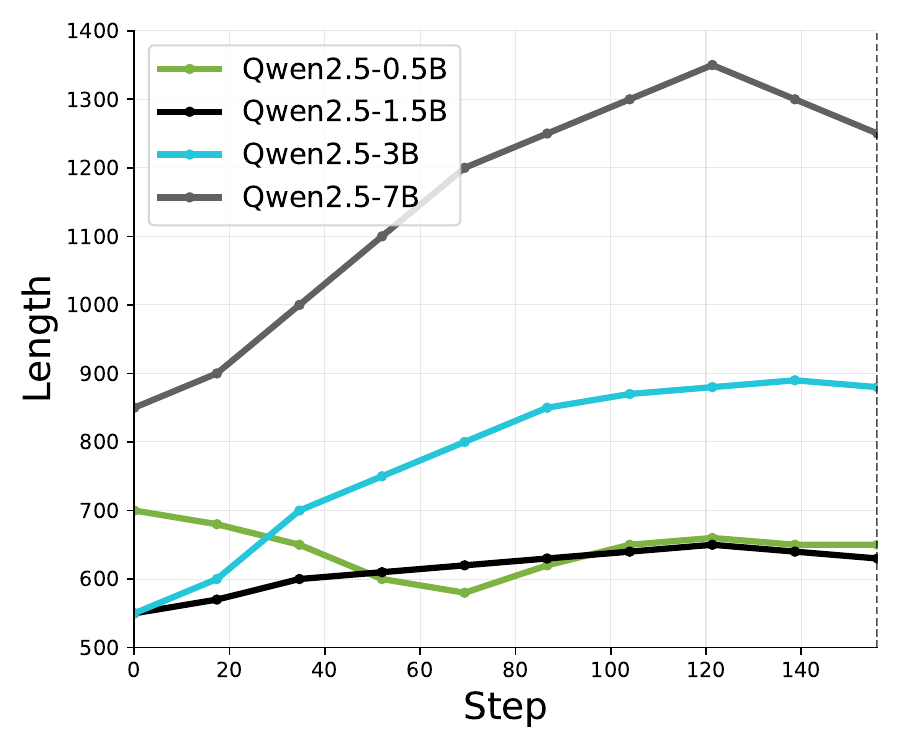}
        \label{fig:sub1}
    \vspace{-10pt}
    \caption{Comparison of average response length over the training steps.}
    \label{fig:wrap-two}
    \vspace{-15pt}
\end{wrapfigure}
with improvements ranging from modest to substantial.
In stark contrast, the Qwen2.5-0.5B model exhibits fundamentally different behaviour under the same training regimes. Across all benchmarks and methods, we observe either only marginal gains or, more concerningly, performance degradation. On Math 500, the 0.5B base model achieves 23.4, but its performance declines when trained with Intuitor, and the model completely collapses when trained with TTRL. Similar patterns are observed across the other benchmarks. To further investigate the cause of model collapse, in Figure \ref{fig:rollouts} we plot the number of correct rollouts at the early stages of training for the base Qwen2.5-0.5B model. As shown, the rollouts generated by the base model often contain no correct outputs. Consequently, majority voting in TTRL produces incorrect pseudo-labels. Training on these erroneous pseudo-labels ultimately leads to model collapse.

\vspace{-5pt}
\subsection{Weaker models do not generate long chain-of-thought (CoT)}
\vspace{-7pt}
One interesting trend we observe is that the reasoning COT length for the stronger model (Qwen2.5-7B) increases as training progresses. As noted in prior work \cite{guo2025deepseek}, longer reasoning chains often include self-reflection (the so-called “Aha moment”). In contrast, the weaker model does not exhibit such long chains. However, generation length alone is not a definitive indicator of improved reasoning. For example, Qwen 0.5B and 1.5B show similar reasoning length, even though the performance is much better for the 1.5B variant.

\vspace{-5pt}
\subsection{Difficulty of training data plays a critical role in label-free RL}
\begin{wrapfigure}{r}{0.35\linewidth}
\vspace{-15pt}
    \centering
    \includegraphics[width=0.99\linewidth]{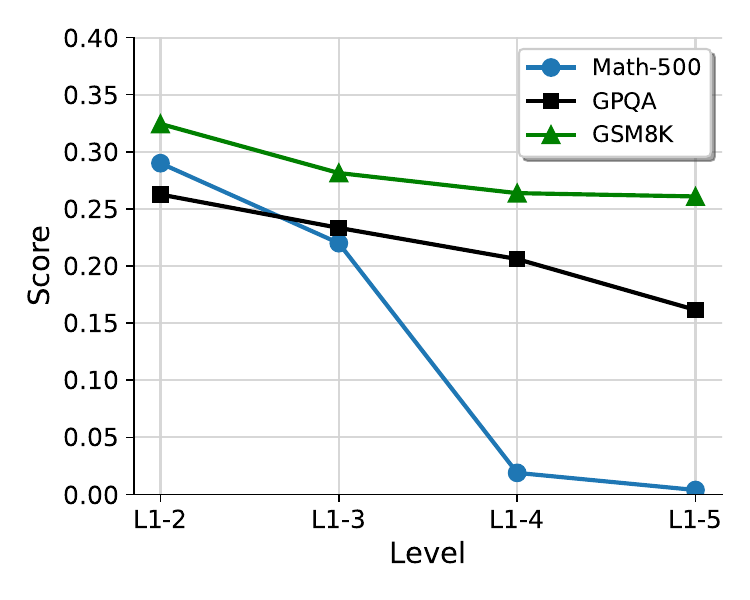}
    \caption{Correct answers generated in early stages of training by Qwen2.5-0.5B.}
    \label{fig:diffculty}
    \vspace{-5pt}
\end{wrapfigure}
To better understand the role of training data distribution, we study the effect of increasing task difficulty on the Qwen2.5-0.5B model. Figure \ref{fig:diffculty} shows performance trends across Math-500, GPQA, and GSM8K as the training data shifts from relatively easier subsets (Level 1-2) to harder or less aligned subsets (Level 1-5).
A clear degradation emerges as the data difficulty increases. For Math-500, the performance drops sharply, with the model nearly collapsing by Level 1-4. GPQA and GSM8K show a similar downward trend, though less steep. This suggests that weak base models are particularly sensitive to data complexity and distributional mismatch. These results highlight an important principle: choosing the right difficulty of training data is critical for effective learning. If the data is too challenging or comes from an unfamiliar distribution, weaker models may fail to generalize and instead suffer from performance degradation. Conversely, aligning training data difficulty with the model’s baseline capacity appears essential for stable improvement

\section{Method}
\vspace{-7pt}
To enhance the reasoning capabilities of language models while learning in a label-free setting, we propose \textbf{Cu}rriculum-guided \textbf{Ma}sked Majority Voting Reinforcement Learning (\textbf{CuMa}). This method utilizes a curriculum learning approach to guide a reinforcement learning process with a majority voting reward signal. By starting with easy samples and gradually increasing the difficulty, the model can learn the reasoning process effectively. Additionally, we introduce a data curation pipeline for generating synthetic data of predefined difficulty levels to stabilize model training and improve reasoning performance. Finally, we employ a reward-masking approach for training in the GRPO setup, where we mask the rewards for samples with no majority prediction. As motivated in Figure \ref{fig:rollouts}, many rollouts, especially at the start of training, generate no majority prediction. Masking the learning signal on such samples ensures that the model does not receive negative feedback from inconclusive examples.

Specifically, we partition an unlabeled dataset $\mathcal{D} = \{x_1, \ldots, x_M\}$ (e.g., math problems) into $K = 5$ difficulty bins, $\mathcal{D}_1, \ldots, \mathcal{D}_K$, where $\mathcal{D}_1$ contains the easiest prompts and $\mathcal{D}_K$ the most challenging. Training proceeds sequentially from $\mathcal{D}_1$ to $\mathcal{D}_K$.
For each bin $\mathcal{D}_k$, we sample a batch of prompts $\{x_i\}_{i=1}^{B}$ where $B$ is the batch size. The model generates multiple candidate solutions $\{y_i^{(j)}\}_{j=1}^{N}$ for each prompt $x_i$, where $N$ is the number of candidate solutions per prompt. We then apply reinforcement learning using a reward signal derived from majority voting on these solutions:
\begin{wraptable}{r}{0.45\textwidth}
\vspace{-5pt}
\centering
\small
\setlength{\tabcolsep}{4pt}
\resizebox{0.99\linewidth}{!}{
\begin{tabular}{l@{\hspace{3pt}}r@{\hspace{3.3pt}}r@{\hspace{3.5pt}}r@{\hspace{3.3pt}}r@{\hspace{3pt}}r}
\toprule
\textbf{Method} & \textbf{Math 500} & \textbf{GPQA} & \textbf{AIME24} & \textbf{GSM8K} & \textbf{LCB} \\
\midrule
\textbf{0.5B Models} & & & & & \\

\quad Base Model & \color{gray}23.4 & \color{gray}28.78 & \color{gray}0.36 & \color{gray}26.38 & \color{gray}0.0 \\
\quad GRPO & \color{gray}33.8 & \color{gray}24.24 & \color{gray}0.6 & \color{gray}25.09 & \color{gray}0.6 \\
\quad Intuitor & {20.8} & \textbf{26.26} & 0.2 & 0.68 & \textbf{0.8} \\
\quad TTRL & 0.0 & 7.57 & 0.0 & 26.38 & 0.0 \\
\rowcolor{table-blue!66}
\quad Ours & \textbf{32.8}	& {22.72}&	\textbf{0.52}	& \textbf{32.9}	& 0.2\\
\midrule

\textbf{1.5B Models} & & & & & \\
\quad Base Model & \color{gray}3.8 & \color{gray}17.17 & \color{gray}0.16 & \color{gray}55.26 & \color{gray}0.07 \\
\quad GRPO & \color{gray}57.4 & \color{gray}23.23 & \color{gray}3.33 & \color{gray}58.90 &\color{gray} 2.42 \\
\quad Intuitor & 47.0 & 22.22 & 1.4 & 44.57 & \textbf{4.85} \\
\quad TTRL & {53.6} & 25.25 & \textbf{3.85} & 58.45 & {2.45} \\
\rowcolor{table-blue!66}
\quad Ours & \textbf{54.2} &	\textbf{25.75}	& {2.49}	& \textbf{59.96} &	3.66 \\
\midrule

\textbf{3B Models} & & & & & \\
\quad Base Model & \color{gray}54.2 & \color{gray}30.80 & \color{gray}3.33 & \color{gray}73.31 & \color{gray}5.20 \\
\quad GRPO & \color{gray}64.4 &\color{gray} 32.32 & \color{gray}5.46 & \color{gray}66.48 & \color{gray}7.65 \\
\quad Intuitor & 59.6 & \textbf{30.30} & 4.01 & 26.56 & 7.83 \\
\quad TTRL & 63.8 & 27.27 & 3.33 & \textbf{74.60} & {7.99} \\
\rowcolor{table-blue!66}
\quad Ours & \textbf{64.4}	& 27.27 & 	\textbf{5.31} &	72.85	& \textbf{8.04}\\
\midrule
\textbf{7B Models} & & & & & \\
\quad Base Model & \color{gray}58.2 & \color{gray}27.77 & \color{gray}6.67 &\color{gray} 81.50 &\color{gray} 4.06 \\
\quad GRPO & \color{gray}73.8 &\color{gray} 37.87 & \color{gray}12.08 & \color{gray} 85.67 & \color{gray}9.91 \\
\quad Intuitor & 72.2 & 27.27 & {12.34} & 78.19 & \textbf{12.5} \\
\quad TTRL & 73.6 & 31.31 & 11.14 & {84.39} & 8.65 \\
\rowcolor{table-blue!66}
\quad Ours & \textbf{74.0}	& \textbf{32.32} &	\textbf{13.33}	& \textbf{84.49}	& 10.31\\ 
\bottomrule
\vspace{-10pt}
\end{tabular}
}
\caption{Performance comparison across different model sizes and methods. Results of existing methods are reproduced in an identical training setup for fair comparison.}
\label{tab:performance_comparison}
\vspace{-15pt}
\end{wraptable}
\begin{equation}
\small
r(x_i, y_i^{(j)}) = \mathbb{I}[y_i^{(j)} = \text{majority\_vote}(\{y_i^{(1)}, \ldots, y_i^{(N)}\})]
\end{equation}
where $\mathbb{I}[\cdot]$ is the indicator function that returns 1 if the condition is true and 0 otherwise. Since small models often struggle with hard samples, this curriculum-based approach allows them to build foundational reasoning skills on easy problems first, which in turn helps them gradually learn to solve more difficult samples.

Another key component of our proposed solution is the reward masking mechanism for samples where $\max_j |\{k : y_i^{(k)} = y_i^{(j)}\}| < 2$, i.e., no majority consensus exists among the $N$ candidates. During early training, small models generate diverse incorrect solutions, creating scenarios with no dominant answer.
Rather than assigning arbitrary rewards to these inconclusive samples, we mask their learning signal entirely:
\begin{equation}
\text{mask}(x_i) = \mathbb{I}\left[\max_j |\{k : y_i^{(k)} = y_i^{(j)}\}| \geq 2 \right]
\end{equation}
This prevents learning from noisy feedback when predictions are uncertain, focusing the RL process on high-confidence samples while avoiding interference from ambiguous cases. The approach is particularly beneficial during initial curriculum phases when the majority consensus reliably indicates solution quality. 

To facilitate further learning from easier samples before being exposed to hard samples, we have curated additional unlabelled samples by using LLM as the data generator, creating synthetic problems at varying difficulty levels to augment the training curriculum. Details on the data curation pipeline are presented in the Appendix.

\section{Results}
\vspace{-5pt}
Table~\ref{tab:performance_comparison} reports the performance of our \textbf{CuMa} method compared to baselines and state-of-the-art approaches across multiple reasoning benchmarks. Overall, \textbf{CuMa} consistently improves reasoning performance, particularly on challenging tasks such as Math 500 and GSM8K, achieving substantial gains over GRPO, Intuitor, and TTRL.  
Our approach is effective across model scales, from 0.5B to 7B parameters, and demonstrates strong generalization across datasets. While performance gains are observed at all model sizes, improvements are more pronounced for smaller models with weaker prior reasoning capabilities. Importantly, we do not observe model collapse on any dataset or scale. We provide additional experiments and implementation details in the Appendix.

\section{Conclusion}
This work systematically explores the limitations of label-free reinforcement learning for enhancing reasoning in large language models, particularly smaller models with limited baseline capabilities. Our analysis reveals that existing methods struggle with smaller models due to insufficient chain-of-thought diversity and sensitivity to training data difficulty, often leading to performance degradation. To address this, we propose \textbf{CuMa}, a curriculum-guided masked majority voting RL approach that leverages progressive difficulty, curated synthetic data, and reward masking to achieve consistent improvements across model sizes (0.5B to 7B) on reasoning benchmarks.

\clearpage
\bibliographystyle{unsrt}
\bibliography{ref}

\appendix
\clearpage
\section{Related Work}
Reinforcement Learning~\citep{sutton1998reinforcement} has become central to improving the reasoning and instruction-following abilities of large language models. The most prominent example, Reinforcement Learning from Human Feedback (RLHF)~\citep{ouyang2022training}, aligns models with human preferences through reward modeling and optimization methods such as Proximal Policy Optimization (PPO)~\citep{schulman2017proximal}. More recently, Large Reasoning Models (LRMs) like DeepSeek-R1~\citep{guo2025deepseek} have highlighted the role of RL in strengthening multi-step reasoning, often by leveraging rule-based rewards. A notable case is GRPO~\citep{shao2024deepseekmath}, which, unlike RLHF’s broad focus on open-domain alignment, explicitly targets mathematical problem solving by encouraging long chain-of-thought (CoT)~\citep{wei2022chain} generation. Considerable attention has since been devoted to improving the robustness and stability of such rule-based RL methods, including GRPO and PPO~\citep{cui2025process,yu2025dapo,liu2025understanding}. Despite these advances, training still relies on supervised data, while inference requires models to extrapolate extended reasoning on unseen tasks. Moreover, effective reward design remains limited to domains with verifiable outcomes—such as math or code—where correctness can be automatically checked~\citep{hu2025open,wei2025swe}.

Beyond verifiable tasks, researchers have investigated self-rewarding~\citep{yuan2025selfrewardinglanguagemodels,prasad2024self} and self-play~\citep{chen2024self} approaches for unlabeled data. However, most of these efforts either emphasize general instruction following~\citep{yuan2025selfrewardinglanguagemodels,chen2024self} or employ preference-based optimization strategies like DPO~\citep{rafailov2023direct} instead of online RL. In parallel, several concurrent directions~\citep{xu2025genius,zhang2025right,zhao2025absolutezeroreinforcedselfplay} explore semi-supervised or reinforcement-inspired reasoning. TTRL \citep{zuo2025ttrl} is one of the latest work in this direction, which use majority voting as a self-derived reward signal, which reduces susceptibility to reward hacking. Complementary findings by \cite{wang2025reinforcement} show that even a single reward-providing example can substantially improve mathematical reasoning, underscoring the potential of lightweight supervision. Intuitor \cite{zhao2025intuitor} replaces external rewards with the model's intrinsic self-certainty—a measure of confidence defined as the average KL divergence from a uniform distribution over the vocabulary.  
In this work, we build upon the concept of majority voting to generate pseudo-labels, similar to TTRL \citep{zuo2025ttrl}, and propose a new framework to address the limitations arising from noisy pseudo-labels.

\section{Data Generation Pipeline}
\label{sec:data_generation}

To support our curriculum-guided reinforcement learning approach, we developed a data curation pipeline to generate synthetic unlabeled datasets with predefined difficulty levels. We leverage an LLM to create diverse prompts, using a structured prompting strategy that emphasizes generating high-quality, varied samples. The prompt explicitly instructs the LLM to produce prompts aligned with a specified difficulty scale (Levels 1 to 5), where each level is defined by example problems provided in the prompt. These examples are carefully selected to represent the reasoning complexity and problem structure characteristic of each difficulty level, ranging from simple arithmetic for Level 1 to advanced multi-step reasoning for Level 5. To ensure robust dataset creation, we generate batches of 25 samples per iteration, allowing for sufficient volume while maintaining computational efficiency.

\begin{wraptable}{r}{0.4\textwidth}
\vspace{-10pt}
\centering
\small
\setlength{\tabcolsep}{4pt}
\resizebox{0.99\linewidth}{!}{
\begin{tabular}{lc}
\toprule
Setting & Performance\\ \midrule
Ours & 32.8 \\
w/o reward masking & 30.7 \\
w/o curated data & 24.5\\ 
w/o curriculum & 20.1 \\
\bottomrule
\end{tabular}
}
\caption{Ablation study}
\label{tab:performance_comparison}
\end{wraptable}
To mitigate bias toward the provided example prompts and promote diversity in the generated dataset, we dynamically refresh the example problems included in each prompting iteration. This dynamic sampling approach ensures that the LLM does not overfit to specific patterns in the provided examples, resulting in a broader range of problem types and structures within each difficulty bin. The generated prompts are then partitioned into $K=5$ difficulty bins, $\mathcal{D}_1, \ldots, \mathcal{D}_K$, based on their alignment with the defined difficulty criteria. This curated dataset is used to train models in our \textbf{CuMa} framework, enabling a progressive learning curriculum that aligns with the model's reasoning capacity and enhances generalization across tasks.

The prompt used for curating our dataset is provided below:

\noindent
\fbox{%
  \begin{minipage}{38.5em}
  \small
  \texttt{%
Data curation prompt: \\
\\
""" \\
You are a math reasoning question generator for LLM training. Generate few high-quality reasoning questions that should be self-contained, promote step-by-step thinking, and not require external knowledge beyond basic facts. \\
Here are the texts:  \\ 
\\
Key requirements: \\
- Do not provide answers, solutions, or reasoning chains. Output only the questions with their difficulty labels.\\
- Vary the questions to cover different sub-topics, including but not limited to ('Algebra', 'Counting \& Probability', 'Geometry', 'Intermediate Algebra', 'Number Theory', 'Prealgebra', 'Precalculus').\\
- Ensure questions are original and engaging.\\
- Include a difficulty level for each question on a scale of 1-5 (1: very easy, basic logic; 5: very hard, multi-step or abstract reasoning).\\
- Target difficulty level: \{target\_level\}. \\
- Examples of level 1 questions: \{level\_1\_examples\}\\
- Examples of level 2 questions: \{level\_2\_examples\}\\
- Examples of level 3 questions: \{level\_3\_examples\}\\
- Examples of level 4 questions: \{level\_4\_examples\}\\
- Examples of level 5 questions: \{level\_5\_examples\}\\
\\
The examples above are for illustration only, and to distinguish between different difficulty levels. Your generated questions must be different from these examples.\\
\\
Output format: \\
- Level \{target\_level\}; Type: [Sub-topic]; [Question text]\\
- Level \{target\_level\}; Type: [Sub-topic]; [Question text]\\
... (repeat for N questions)\\
\\
Generate exactly N questions of Level \{target\_level\}.\\
"""%
  }
  \end{minipage}%
}

\section{Experiments}
\subsection{Implementation details}
\vspace{-7pt}
To implement our \textbf{CuMa} method, we apply Group Relative Policy Optimization independently across each benchmark, adapting the approach outlined in \cite{guo2025deepseek}. We utilize a cosine learning rate schedule with a peak learning rate of $3 \times 10^{-6}$ and employ the AdamW optimizer for policy optimization. For each training prompt, we generate 8 candidate responses using a temperature of 0.6 for majority voting to estimate pseudo-labels. The maximum generation length is capped at 3,072 tokens for all models. The number of training episodes is set to 1. All experiments were conducted on a cluster of 4 NVIDIA H100 80GB GPUs. 
\subsection{Ablation study}
To evaluate the contributions of each component in our \textbf{CuMa} framework, we conducted an ablation study on the Math 500 benchmark using the Qwen2.5-0.5B model, with results summarized in Table~\ref{tab:performance_comparison}. Our full method achieves a performance score of 32.8. Removing the reward-masking mechanism, which excludes samples without a majority consensus, reduces the score to 30.7, indicating that masking inconclusive rollouts is crucial for stabilizing training and preventing learning from noisy feedback. Omitting the curated synthetic data generated by our data curation pipeline further degrades performance to 24.5, highlighting the importance of diverse, difficulty-controlled samples in enhancing reasoning capabilities. Finally, excluding the curriculum learning strategy, which progressively introduces harder problems, results in a significant drop to 20.1, underscoring the necessity of aligning training data difficulty with the model's capacity to avoid performance degradation. These results confirm that each component—reward masking, curated data, and curriculum learning—plays a critical role in the effectiveness of \textbf{CuMa} for label-free reinforcement learning across varying model sizes.

\end{document}